\documentclass[conference]{IEEEtran}
\IEEEoverridecommandlockouts    

\usepackage{multirow}
\usepackage{lipsum}
\usepackage{amsmath}
\usepackage{amssymb}
\usepackage[ruled,vlined]{algorithm2e}
\usepackage{graphicx}
\graphicspath{{./Figures/}}
\usepackage{orcidlink}
\usepackage{booktabs}
\usepackage{xurl}

\title{\vspace{0.20in}\LARGE \bf Learning to See Like Humans: Gaze-Aligned Cycling Safety Prediction}

\author{
	\parbox{\textwidth}{%
		\centering
        Luís Maria Perdigão\,\orcidlink{0009-0007-5355-1702}$^{1}$, 
        Miguel Costa\,\orcidlink{0000-0003-0860-7002}$^{2}$, 
        Carlos Santiago\,\orcidlink{0000-0002-4737-0020}$^{1}$, 
        Manuel Marques\,\orcidlink{0000-0003-0532-1869}$^{1}$%
	}%
    \thanks{$^{1}$Institute for Systems and Robotics (ISR-Lisboa), Instituto Superior Técnico, Lisbon, Portugal. 
        {\tt\small luis.perdigao@tecnico.ulisboa.pt}}
    \thanks{$^{2}$Technical University of Denmark, Kongens Lyngby, Denmark.}
}

\begin{document}
	
	\maketitle
	\thispagestyle{empty}
	\pagestyle{empty}
	
	\begin{abstract}
		Cycling delivers significant public-health and environmental benefits, yet its uptake in cities is often limited by perceived safety. When street environments appear unsafe, individuals are less likely to cycle, making perception a key barrier to adoption. Recent work has shown that pairwise comparisons of street-view images provide a scalable way to learn subjective safety judgments. However, existing approaches do not explicitly model human visual attention, which plays a central role in how humans perceive safety. We propose an Eye-Tracking–Guided Perceived Cycling Safety framework (EG-PCS) that integrates gaze data into a pairwise learning pipeline based on vision transformers. By supervising the model’s attention mechanism with eye-tracking signals, we encourage alignment between learned attention maps and human fixation patterns. Experiments show that gaze-guided models achieve similar ranking performance compared to state-of-the-art approaches while producing attention maps that more accurately reflect human visual attention behavior. Our results demonstrate that incorporating eye-tracking information enhances both predictive accuracy and interpretability in perception-based urban analytics.
	\end{abstract}
	

    \section{Introduction}
    \label{sec:introduction}
    
    Cycling is widely promoted as a sustainable urban transport mode because it supports public health and more liveable cities \cite{Oja2011,PucherBuehler2012}.
    Despite strong policy efforts in many cities to promote cycling and increase modal share, cycling levels often remain below target goals, in part because many people perceive cycling as stressful or dangerous in traffic \cite{DillCarr2003,ParkinWardmanPage2008}.
    This makes perceived safety a central barrier: even when cycling is feasible, riders may avoid it if the built environment \emph{looks} unsafe \cite{WintersTeschke2010,vonStlpnagel2020}.
    
    However, understanding perceived cycling safety at scale remains challenging. Traditional approaches such as surveys, interviews, or controlled field studies provide rich qualitative insight. This process is slow, expensive, and difficult to generalize and transfer across cities and contexts. As a result, recent work has increasingly tried to leverage street-level imagery and computer vision to assess cycling conditions over large geographic areas \cite{ItoBiljecki2021,Ye2024}.
    

    One of the most successful approaches for learning image-level perceptions relies on pairwise comparisons between images. Urban perception studies \cite{Naik2014, Dubey2016} and, more recently, \cite{MiguelCosta2025} in the context of cycling, introduced dedicated pairwise learning frameworks. Based on hundreds of survey responses, these works train deep learning models to predict perceptual attributes of urban environments, such as beauty, liveliness, wealth, or perception of cycling safety. While such approaches enable scalable ranking of environments, they do not explicitly ensure that semantically meaningful regions of images---such as people, sidewalks, cars, or buildings \cite{Costa2019}---are effectively leveraged in the estimation process. Since perceived cycling safety arises from human visual assessment, its estimation should be aligned with human visual behavior. 


    
    \begin{figure}[t]
        \centering
        \includegraphics[width=\linewidth]{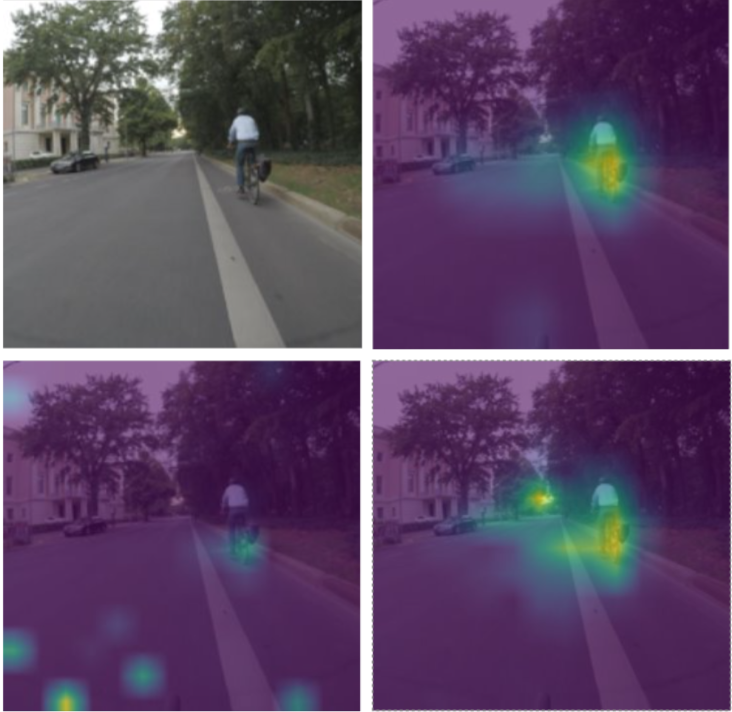}
        \caption{Example of human vs.\ model attention on a street scene with a bike lane and a cyclist riding ahead. \textbf{Top-left:} the original input image. \textbf{Top-right:} human attention (eye-tracking saliency) while judging cycling safety, concentrating on the cyclist and the riding corridor. \textbf{Bottom-left:} transformer attention from a model trained \emph{without} gaze supervision, which places attention on regions that are less consistent with human fixations (eg. sky attention on the top-left corner). \textbf{Bottom-right:} transformer attention from EG-PCS-Net trained \emph{with} gaze supervision, showing substantially closer agreement with human attention and additionally highlighting the road vanishing point as a task-relevant region for perceived safety.}
        \label{fig:intro_alignment}
    \end{figure}
    
    In this work, we build on previous pairwise comparison frameworks and incorporate eye-tracking data to guide the attention mechanism of vision transformers \cite{ViT2020} to enhance both predictive performance and interpretability. As illustrated in Fig.~\ref{fig:intro_alignment}, models trained with gaze supervision generate attention maps that more closely align with human fixation patterns. Thus, our main contributions are:
    
    \begin{itemize}
        \item We introduce EG-PCS-Net (Eye-Tracking–Guided Perceived Cycling Safety), a Siamese Vision Transformer framework that jointly optimizes pairwise classification, ranking, and attention–gaze alignment, explicitly encouraging correspondence between model self-attention and human fixation in perceived cycling safety judgments.
    
        \item We conduct a systematic evaluation of multiple state-of-the-art gaze-integration strategies—including attention alignment, patch-selection masking, and gaze-feature injection—across three pretrained transformer backbones, showing that our explicit attention supervision substantially improves attention–gaze agreement while maintaining competitive predictive performance.
    
        \item We make available a dataset of 13,623 labeled pairwise street-view comparisons for perceived cycling safety, of which 1,495 comparisons include eye-tracking data.
    \end{itemize}
    
    \section{Related Work}
    \label{sec:relatedwork}
    
    \subsection{Perception of Safety}
    Perceived cycling safety is repeatedly reported as a key factor shaping cycling uptake and route choice, complementing (and sometimes diverging from) objective safety measured through crashes and conflicts \cite{DillCarr2003,ParkinWardmanPage2008,WintersTeschke2010,vonStlpnagel2020,costa2025bridging}. Early research often relied on qualitative surveys and interviews (in situ or post-ride) to identify environmental elements that increase stress or fear \cite{MiguelCosta2025}. While these methods provide key insights and detailed explanations on why people perceive certain environments as more dangerous than others, they are typically costly, time-consuming, and difficult to scale or replicate across cities and infrastructure typologies.
    
    More recent work has used new measurement technologies to capture  more quantitatively responses, including wearable sensors, cycling videos, mental maps, virtual reality, street-view-style imagery, and eye-tracking \cite{CyclingNet,Zhang2024,Zeuwts2023,MiguelCosta2025}. A major enabler is the rapid growth of large-scale street-level imagery, with near-global coverage becoming available through platforms such as Google Street View, Mapillary and similar sources, which makes it possible to study cycling environments across many cities without collecting imagery manually \cite{Dubey2016,ItoBiljecki2021,MiguelCosta2025}. In parallel, planning practice has long used indices that approximate cycling comfort or perceived risk (e.g., Bicycle Stress Level and Level of Traffic Stress), but these often require manual annotation and hand-crafted rules. To reduce manual effort, computer-vision pipelines have been proposed to compute “bikeability” or safety-related indicators from street-view imagery by first extracting features (objects, layout, semantics). These can then be mapped to bikeability or perception scores \cite{Song2018,ItoBiljecki2021, Ramrez2021} or used to extract correlations of features that lead to environments being perceived as safe or unsafe \cite{Rita2023London}.
    While these approaches can provide useful indicators of cycling attractiveness, they rely on predefined features that may not fully capture human perception, which ultimately drives the decision to cycle.
    
    
    Learning perceived safety \emph{directly} from images using human judgments avoids part of this hand-design problem and can further improve scalability by letting models learn relevant cues from pixels rather than from pre-defined indicators. 
    Pairwise data can be turned into global rankings using rating systems such as TrueSkill \cite{Herbrich2006TrueSkill} or via optimization-based estimators \cite{PereiraCorreiaCabraldaCosta2019}. Compared to single-image scoring, comparisons are often an easier task (“which looks safer?”) and can reduce noise when perceptions vary across people and contexts \cite{Dittrich2005}. Dubey et al. \cite{Dubey2016} advanced earlier work by scaling up pairwise datasets and training deep models directly on comparisons, improving generalization from visual content rather than relying only on post-processed image scores and \cite{MiguelCosta2025} uses a similar framework in a cycling safety scenario. 

    
    Still, across the literature there are methods that are strong at \emph{localizing} safer/unsafe areas \cite{MiguelCosta2025, Ito2021, PereiraCorreiaCabraldaCosta2019} through scalable scoring and methods that are strong at \emph{explaining} the design cues that drive those perceptions (\cite{Rita2023London, Costa2019}). However, the methods lack in trying to do both at the same time. Pairwise deep models can rank environments effectively, but the “why” behind a score is often less explicit than in feature-based studies that target specific built-environment variables. This gap motivates extending pairwise cycling-safety learning with vision-transformer backbones.

    \subsection{Vision Transformers and Attention Guidance}
    
    Vision Transformers (ViTs) \cite{ViT2020} represent an image as a sequence of patches and allow every patch to directly interact with every other patch through self-attention. Unlike convolutional networks, which typically require separate post-hoc methods to explain their predictions, ViTs naturally produce attention weights that indicate how different image regions influence one another. These attention patterns provide an explicit signal that can be inspected to understand which parts of the image contribute to the final decision.
    
    
    
    Guided-attention approaches take a complementary step by shaping attention during training rather than only analyzing it afterward. Earlier work introduced mechanisms where attention maps are explicitly regularized or refined during learning \cite{Li2018TellMeWhere}. More recently, gaze-guided transformers (\cite{Zhuang2025, Hu2025, Koorathota2024}) have incorporated expert eye-tracking data as an inductive bias. For example, EG-ViT masks patch tokens outside gaze-relevant regions to discourage shortcut learning \cite{Ma2023EGViT}, while Chen et al.\ integrate gaze features directly inside transformer blocks through a Gaze Information Injector (GII) \cite{Chen2026GIIViT}. Although these methods were originally developed for medical imaging tasks, their underlying principle (using human gaze to guide representation learning) is general and can be transferred to other domains, including subjective perception modeling such as cycling safety assessment.
    
    These approaches suggest a broader perspective: if attention is to be used as an interpretability signal, it may benefit from supervision that aligns it with human visual behavior. In this context, incorporating gaze during training does not merely provide additional information for prediction, but it can also increase confidence that the model’s internal attention maps reflect what humans actually consider when making judgments. This perspective is particularly relevant when modeling subjective constructs such as perceived cycling safety, where interpretability is central for urban planners. From a planning perspective, it is valuable not only to localize infrastructure associated with lower perceived safety, but also to identify the specific visual cues that drive those perceptions.
    
    \section{Proposed Method}
    \label{sec:method}
    
    \subsection{Data: Pairwise Comparisons and Eye-Tracking Subset}
    \label{sec:data}
    
    Our experiments use the pairwise Cycling Safety Perception (PCS) dataset introduced by Costa et al.\ \cite{MiguelCosta2025}, alongside an eye-tracking subset we collected under the same pairwise protocol. We denote the dataset by $\mathcal{D} = \{(I_L, I_R, G_L, G_R, y)\}$, where each sample consists of a pair of street-view images $(I_L, I_R)$, the corresponding gaze saliency maps $(G_L, G_R)$ when available, and a label $y$ indicating which scene is perceived as safer to cycle (or both scenes are perceived equally safe). Although the full PCS dataset includes ties, the eye-tracking subset does not provide tie annotations. For consistency across all gaze-related experiments, we restrict learning and evaluation to non-tie comparisons only. The label is defined as $y \in \{-1, 1\}$, where $y=-1$ denotes that the left image is perceived as safer and $y=1$ denotes that the right image is perceived as safer.
        
        
    
    The PCS dataset $\mathcal{D}$ comprises pairwise comparisons collected across multiple cities (e.g., Paris, London, Barcelona, Munich, and Berlin). Among these, the Berlin subset contains 5,907 non-tie comparisons, of which 999 include eye-tracking annotations. The largest portion of eye-tracking data is available for the Berlin subset, therefore, we focus on the Berlin subset for gaze-guided training and benchmarking, as it provides the only setting where attention alignment can be learned and evaluated in a stable manner.



\begin{figure}[t]
        \centering
        \includegraphics[width=\linewidth]{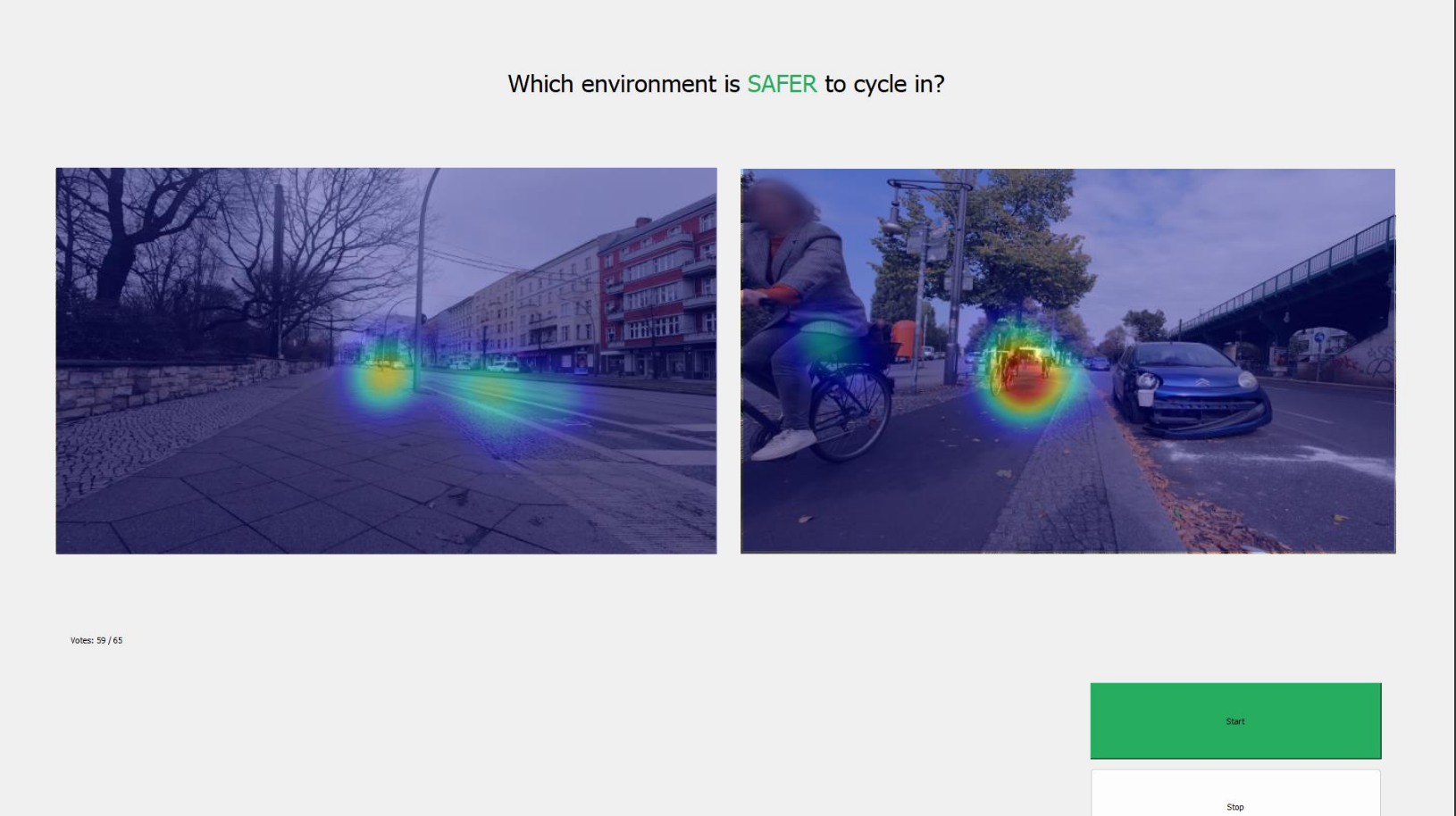}
         \caption{Layout used in the eye-tracking experiment. Participants were asked to choose which environment appeared safer to cycle in. The displayed images include an overlaid gaze-based heatmap, illustrating the spatial distribution of visual fixations during the decision process.}
        \label{fig:trial_ui}
    \end{figure}

    Comparisons were answered by 249 different respondents, from which, 23 used eye-tracking technology. Eye movements were recorded using a Tobii eye tracker at 60\,Hz, synchronized with a custom Python/PyQt interface to control stimulus timing. Each participant performed a calibration and then completed 65 trials. In each trial, two Berlin street-view images were shown side-by-side and the participant was asked to select the scene he/she thought was safer to cycle in. Figure~\ref{fig:trial_ui} shows an example of the interface used by respondents.

    Gaze recordings were processed in OGAMA \cite{Vosskuhler2008OGAMA}, which outputs fixation events (screen coordinates and durations) using a dispersion-threshold, moving-window fixation detector \cite{SalvucciGoldberg2000}. To obtain a continuous gaze saliency map, we follow standard saliency-map construction used in visual-attention benchmarking: a fixation map is smoothed with an isotropic Gaussian \cite{LeMeur2013}. This follows the common assumption that one degree of visual angle \cite{Bylinskii2017} approximates foveal scale and that a fixation can be modeled as a Gaussian whose spread depends on viewing distance and display geometry.
    
    Let $\mathcal{G}=\{(x_n,y_n,w_n)\}_{n=1}^{N_f}$ be the fixation list for one image, where $(x_n,y_n)$ are pixel coordinates and $w_n$ is the fixation duration (or unit weight if durations are not used). The discrete fixation map is
    \begin{equation}
    f(x,y) \;=\; \sum_{n=1}^{N_f} w_n \,\mathbf{1}\!\left[(x,y)=(x_n,y_n)\right],
    \end{equation}
    where $\mathbf{1}[\cdot]$ is the indicator function (1 if the condition holds, 0 otherwise). The gaze smoothed saliency map is then
    \begin{equation}
    G(x,y) \;=\; (f * G_{\sigma})(x,y),
    \label{eq:gaze_conv}
    \end{equation}
    where $G_{\sigma}$ denotes a 2D Gaussian smoothing filter with standard deviation $\sigma$, and $*$ denotes 2D convolution, with $\sigma$ set to $1^\circ$ of visual angle \cite{LeMeur2013,Bylinskii2017}.

    The released metadata does not include viewing distance or physical monitor size. We therefore approximate the viewing distance as $d=50$\,cm and assume a typical 24'' monitor at $1920\times1200$. 
    

    \subsection{EG-PCS-Net Architecture}
    \label{sec:arch}

    \begin{figure*}[t]
        \centering
        \includegraphics[width=\linewidth]{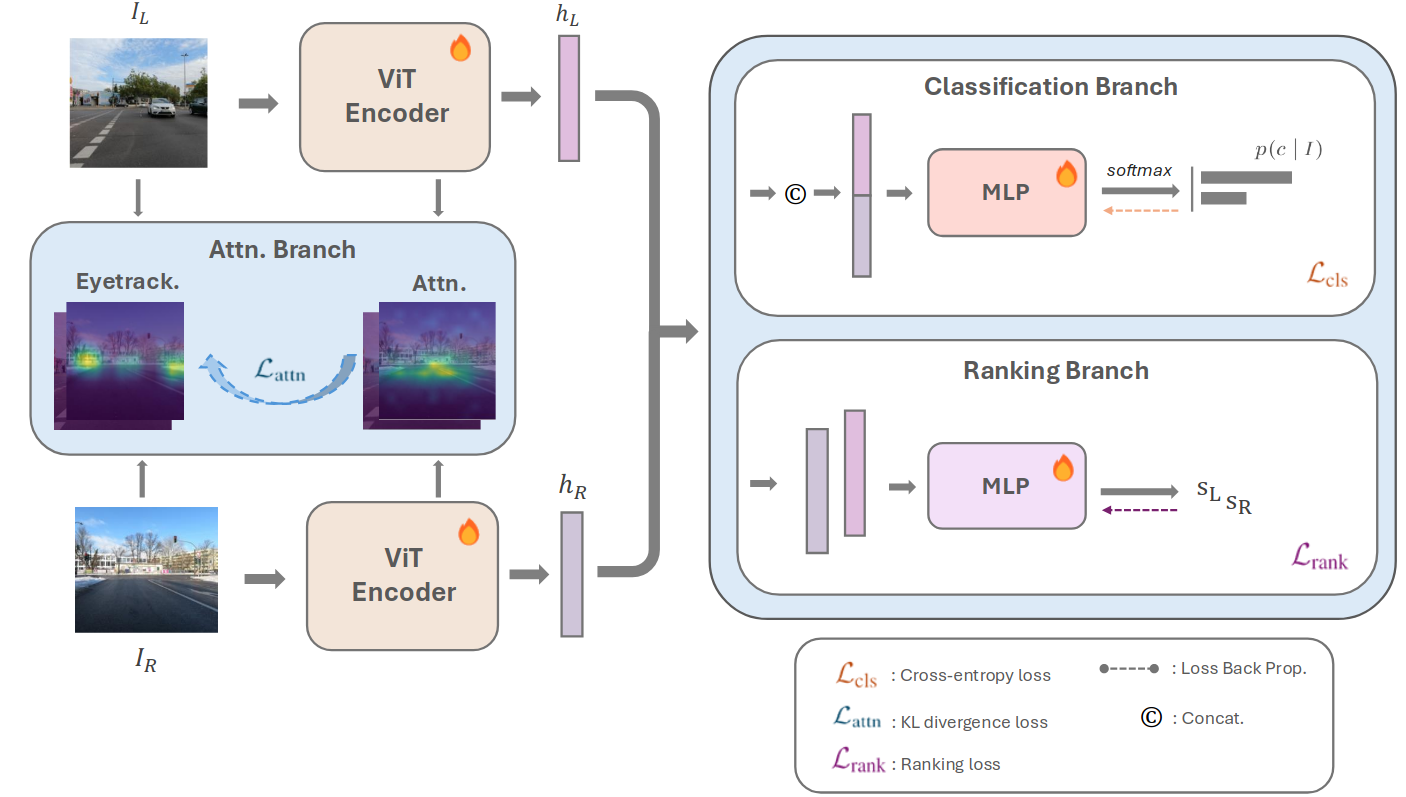}
        \caption{EG-PCS-Net architecture. Two input images, $I_L$ (top-left) and $I_R$ (bottom-left), are processed by a shared Vision Transformer encoder (Siamese backbone). Each image is tokenized and passed through $L$ transformer layers, producing contextualized \texttt{CLS} representations $h_L$ and $h_R$. On the right, the architecture branches into two prediction heads. In the classification branch (top-right), the concatenated representation $[h_L;h_R]$ is fed to an Multilayer Perceptron (MLP) followed by a softmax layer that produces the class probabilities $p(c \mid I)$, optimized via the cross-entropy loss $\mathcal{L}_{\text{cls}}$. In parallel, the ranking branch (bottom-right) maps each descriptor independently to scalar safety scores $s_L$ and $s_R$ through a shared MLP, trained with the margin-based ranking loss $\mathcal{L}_{\text{rank}}$. In the middle-left, the attention branch extracts class-to-patch attention maps from the transformer and compares them with gaze-derived saliency maps using a KL-divergence loss $\mathcal{L}_{\text{attn}}$. All three objectives are jointly optimized.}
        \label{fig:egpcsnet_arch}
    \end{figure*}
    
    We propose a new architecture that augments the original perceived safety-ranking framework with an explicit attention branch. The goal of this extension is not only to achieve the strong ranking and classification performance, but also to provide a higher degree of confidence when interpretability methods are applied. By explicitly extracting and structuring transformer attention signals, EG-PCS-Net enables systematic analysis of which image regions contribute to perceived safety decisions. As illustrated in Fig.~\ref{fig:egpcsnet_arch}, the architecture integrates prediction (classification and ranking) and attention extraction within a unified Siamese Vision Transformer backbone.
    
    Each image $I \in \mathbb{R}^{H \times W \times C}$ is split into non-overlapping $P \times P$ patches, giving
    $N=\frac{HW}{P^2}$ patch tokens. After patch embedding and positional encoding, a learnable class token
    (\texttt{CLS}) is prepended, yielding a token sequence $T=\{t_i\}_{i=1}^{N+1}$, where $t_1$ is the \texttt{CLS}
    token and $t_2,\dots,t_{N+1}$ are patch tokens.

    The sequence $T$ is processed by $L$ transformer encoder layers. At each layer $\ell \in \{1,\dots,L\}$, the model produces contextualized token representations
    \begin{equation}
    z^{(\ell)} \in \mathbb{R}^{(N+1)\times D},
    \end{equation}
    so that information is progressively aggregated across layers. The final-layer output $z^{(L)}$ contains the fully contextualized tokens. In particular, the \texttt{CLS} token integrates information from all patches through self-attention and serves as a global descriptor of the image. For the left and right images, the global descriptors are obtained as the first row (corresponding to the \texttt{CLS} token) of the final-layer outputs:
    \begin{equation}
    h_L = z^{(L)}_{L}[1, :] \in \mathbb{R}^D, \qquad h_R = z^{(L)}_{R}[1, :] \in \mathbb{R}^D,
    \end{equation}
    which are used in the prediction branches.

    The network branches into three components: (i) a classification branch that predicts which side is safer, (ii) a ranking branch that assigns a continuous safety score to each image, and (iii) an attention branch that enforces human-like attention.

    \subsubsection{\textbf{Classification Branch}}
    
    For the pairwise decision, we extract the final \texttt{CLS} representations of each image, $h_L$ and $h_R$, and form a joint representation by concatenation:
    \begin{equation}
    h_{LR} = [\,h_L;\,h_R\,] \in \mathbb{R}^{2D}.
    \end{equation}
    This vector is passed through three fully connected layers followed by a softmax layer that outputs class probabilities
    $p_c$ for $c \in \{-1,+1\}$. The classification loss is the cross-entropy:
    \begin{equation}
    \mathcal{L}_{\text{cls}}
    =
    -\sum_{c \in \{-1,+1\}}
    \mathbf{1}(y=c)\,\log p_c,
    \label{eq:clsloss_final}
    \end{equation}
    where $y$ is the ground-truth label (Sec. \ref{sec:data}).
   
    \subsubsection{\textbf{Ranking Branch}}
    In parallel, the ranking branch learns a continuous perceived safety score for each image. A shared scoring function
    $f(\cdot)$ maps each \texttt{CLS} representation to a scalar:
    \begin{equation}
    s_L = f(h_L), \qquad s_R = f(h_R),
    \end{equation}
    where larger values correspond to higher perceived cycling safety. Consistency between score ordering and ground-truth preference $y$ is enforced through a margin-based ranking loss:
    \begin{equation}
    \mathcal{L}_{\text{rank}}
    =
    \max\!\left(0,\; \gamma - y\,(s_R - s_L)\right),
    \label{eq:rankloss_final}
    \end{equation}
    where $\gamma>0$ is a margin hyperparameter.

    \subsubsection{\textbf{Attention branch}}
    \label{sec:attn_branch}
    
    Beyond prediction, EG-PCS-Net extracts spatial attention maps to provide model interpretability. For each layer $\ell$ and head $h \in \{1,\dots,H\}$, the query and key matrices $Q^{(\ell,h)}, K^{(\ell,h)} \in \mathbb{R}^{(N+1) \times D'}$ are used to compute the self-attention weights:
    \begin{equation}
    A^{(\ell,h)} = \mathrm{softmax}\left( \frac{Q^{(\ell,h)} {K^{(\ell,h)}}^\top}{\sqrt{D'}} \right) \in \mathbb{R}^{(N+1) \times (N+1)},
    \end{equation}
    where $D' = D/H$ is the head dimension. These weights represent the interactions between the full token sequence, including the \texttt{CLS} token and the $N$ patch tokens. To obtain a single map per layer, we average $A^{(\ell,h)}$ across all $H$ heads to yield $\bar{A}^{(\ell)}$. Two extraction strategies are considered.
    
    \textbf{Raw attention} specifically uses the first row (corresponding to the \texttt{CLS} token, $t_1$) from the averaged attention matrix $\bar{A}^{(L)}$ of the final layer $L$. Since the \texttt{CLS} token serves as the global image descriptor for the classification and ranking branches, its attention distribution over the $N$ patches offers a natural spatial signal reflecting the regions the model prioritizes for its decision.
    
    \textbf{Attention rollout} composes attention across all layers by recursively multiplying residual-adjusted attention matrices:
    \begin{equation}
    R = \tilde{A}^{(1)} \tilde{A}^{(2)} \cdots \tilde{A}^{(L)} \in \mathbb{R}^{(N+1) \times (N+1)},
    \end{equation}
    where $\tilde{A}^{(\ell)}$ denotes the head-averaged attention $\bar{A}^{(\ell)}$ with identity added to account for residual connections, followed by row-normalization. The row in $R$ corresponding to the \texttt{CLS} token provides a multi-layer summary of information flow from patches to the global representation. 
    
    In both cases, the class-to-patch weights corresponding to the \texttt{CLS} token define model attention maps $M_L, M_R \in \mathbb{R}^{N}$ for the left and right images, which are used in the attention branch. 
    
    When eye-tracking is available, the model aligns these attention maps with human gaze. Let $G_L$ and $G_R$ denote the gaze saliency maps described in Sec.~\ref{sec:data}. Both gaze and model attention maps are represented as discrete distributions over the $N$ patch locations. The gaze maps are normalized to obtain $\hat{G}_L$ and $\hat{G}_R$. Alignment is enforced through a KL-divergence loss:
    \begin{equation}
    \mathcal{L}_{\text{attn}}
    =
    \frac{1}{2}
    \left(
    \sum_{i=1}^{N}
    \hat{G}_{L,i} \log\frac{\hat{G}_{L,i}}{M_{L,i}}
    +
    \sum_{i=1}^{N}
    \hat{G}_{R,i} \log\frac{\hat{G}_{R,i}}{M_{R,i}}
    \right).
    \label{eq:gazeloss_final}
    \end{equation}
    
    When alignment is computed using \emph{Raw} attention, the loss depends only on the final-layer matrix $\bar{A}^{(L)}$, and gradients primarily affect the last encoder layer. When using \emph{Rollout}, the loss depends on the composed matrix $R$, and gradients propagate through attention matrices across all transformer layers. Accordingly, EG-PCS-Net (Raw) supervises final-layer attention, whereas EG-PCS-Net (Rollout) supervises attention throughout the encoder depth.
    
    The full objective combines classification, ranking, and (when available) gaze alignment:
    \begin{equation}
    \mathcal{L}
    =
    \mathcal{L}_{\text{cls}}
    +
    \lambda_{\text{rank}}\,\mathcal{L}_{\text{rank}}
    +
    \lambda_{\text{gaze}}\,\mathbf{1}[\texttt{has\_gaze}]\,\mathcal{L}_{\text{attn}},
    \label{eq:totalloss_final}
    \end{equation}
    where $\lambda_{\text{rank}}$ and $\lambda_{\text{gaze}}$ control the contribution of ranking and alignment. Together, these three objectives allow EG-PCS-Net to (i) make accurate pairwise safety decisions, (ii) learn a globally meaningful safety ranking over images, and (iii) produce attention maps that can be quantitatively and qualitatively compared to human visual behavior.

    \section{Experiments}
    \label{sec:experiments}
    
    All experiments are conducted using the dataset $\mathcal{D}$ introduced in Sec.~\ref{sec:data}. The dataset is randomly split into 70\%/ 10\%/ 20\% for training/validation/test, at the level of pairwise comparisons. All experiments are implemented in Python using PyTorch~2 \cite{Paszke2019PyTorch} and run on a single NVIDIA GeForce GTX 1080~Ti GPU. The effective batch size is 128 pairwise comparisons. AdamW \cite{Loshchilov2019AdamW} is used as optimizer. Training uses a warmup--cosine learning-rate schedule, combining linear warmup \cite{Goyal2017LargeBatch} with cosine annealing \cite{Loshchilov2016SGDR}.  Early stopping is applied when the validation loss does not improve for 3 consecutive epochs. Code, data-processing scripts and datasets are publicly available\footnote{\url{https://github.com/sipg-isr/Eyetracking-cycling-safety-perception}}.
    
    Across all settings, model selection and hyperparameter tuning are performed on the validation split. This includes the ranking margin $\gamma$ (which controls the desired separation between safer and less-safe scenes in score space), the number of transformer layers unfrozen during fine-tuning, and the relative weights assigned to the classification, ranking, and gaze-related objectives. For methods adapted from prior work, method-specific hyperparameters are set according to the original papers to preserve the intended training dynamics and enable fair comparison.

    
    \subsection{Backbones and Gaze-Integration Modes}
    
    We evaluate three popular pretrained Vision Transformer backbones---DINOv3 ViT-B/14 \cite{simeoni2025dinov3}, DeiT III ViT-B/16 \cite{Touvron2022DeiTIR}, and CLIP ViT-B/16 \cite{radford2021learning}---each plugged into the EG-PCS-Net architecture described in Sec.~\ref{sec:arch}. 
    
    To analyze the effect of gaze supervision, we compare four gaze-integration modes to a baseline:
    
    \begin{itemize}
        \item \textbf{Baseline}: EG-PCS-Net trained without gaze, optimizing only the classification 
        and ranking objectives.
        
        \item \textbf{EG-PCS-Net (Raw)} and \textbf{EG-PCS-Net (Rollout)}: our attention-alignment variants, which introduce the KL-based attention–gaze loss described in Sec.~\ref{sec:arch}. 
        
        \item \textbf{EGViT} \cite{Ma2023EGViT}: a gaze-guided patch-selection strategy adapted to the pairwise cycling-safety setting.
        
        \item \textbf{GII injection} \cite{Chen2026GIIViT}: a gaze-feature injection strategy adapted to the pairwise architecture.
    \end{itemize}
    
    Descriptions of EGViT and GII-style injection are provided in Sec.~\ref{sec:relatedwork}.

    \subsection{Quantitative Analysis}
    
    \begin{table*}[ht]
    \centering
    \caption{Average Accuracy (Rank / Class) with 95\% confidence intervals over tested seeds.}
    \label{tab:gaze_modes_final}
    \begin{tabular}{llccc}
    \toprule
    Method & Attention Mode & DINOv3 & DeiT III & CLIP \\
    \midrule
    Baseline & -- & 74.73$_{{\pm0.59}}$/74.06$_{{\pm0.84}}$ & 73.84$_{{\pm0.84}}$/73.20$_{{\pm1.14}}$ & 73.34$_{{\pm0.56}}$/72.11$_{{\pm1.78}}$ \\
    GII injection \cite{Chen2026GIIViT} & -- & \textbf{74.88}$_{{\pm0.83}}$/74.55$_{{\pm0.75}}$ & 73.29$_{{\pm0.97}}$/72.82$_{{\pm1.18}}$ & \textbf{73.36}$_{{\pm0.74}}$/\textbf{73.42}$_{{\pm0.76}}$ \\
    EGViT \cite{Ma2023EGViT} & -- & 74.41$_{{\pm0.57}}$/74.21$_{{\pm0.70}}$ & 73.94$_{{\pm0.69}}$/73.35$_{{\pm0.78}}$ & 73.03$_{{\pm0.96}}$/72.15$_{{\pm1.42}}$ \\
    EG-PCS-Net (Ours) & Raw & 74.66$_{{\pm0.60}}$/\textbf{74.64}$_{{\pm0.52}}$ & 73.78$_{{\pm0.72}}$/73.52$_{{\pm0.84}}$ & 73.32$_{{\pm0.82}}$/73.06$_{{\pm0.79}}$ \\
    EG-PCS-Net (Ours) & Rollout & 74.32$_{{\pm0.51}}$/74.60$_{{\pm0.56}}$ & \textbf{74.20}$_{{\pm0.88}}$/\textbf{73.93}$_{{\pm0.93}}$ & 72.71$_{{\pm0.65}}$/71.68$_{{\pm1.50}}$ \\
    \bottomrule
    \end{tabular}
    \end{table*}

    We evaluate each backbone and gaze mode along two complementary points of view. First, we measure predictive performance using pairwise \emph{classification accuracy} (left vs.\ right safer) and \emph{ranking accuracy} (whether the learned scores satisfy the correct ordering). 
    
    Second, we evaluate whether introducing gaze supervision improves interpretability by making model attention more human-like. Concretely, we compare model attention maps against the gaze saliency maps using standard saliency-benchmarking metrics \cite{Bylinskii2017}. The reported metrics include Area Under the ROC Curve (AUC), Normalized Scanpath Saliency (NSS), Pearson’s Correlation Coefficient (CC), Earth Mover’s Distance (EMD), Similarity / histogram intersection (SIM), Kullback--Leibler divergence (KL), and Information Gain (IG). This evaluation is designed to answer a specific question: even if predictive accuracy is similar, does gaze-guided training move the model’s self-attention closer to human fixation behavior?

    \subsubsection{Predictive Performance}
    
    Table~\ref{tab:gaze_modes_final} reports the predictive performance for each Backbone/Method combination. Each configuration was evaluated over 10 independent runs with different random seeds. The results are presented as mean accuracy with 95\% confidence intervals. Two main trends emerge. First, introducing gaze does not harm predictive performance: gaze-guided variants remain competitive with the baseline across backbones. Second, the effect depends on how gaze is incorporated and on the backbone. For DeiT III, our method (Rollout) yields the strongest average accuracy, suggesting that rollout-based aggregation may provide a more stable training signal than a single-layer raw attention map in this backbone. For DINOv3, differences between gaze modes are modest, indicating that strong self-supervised pretraining may already produce robust representations where gaze primarily acts as a mild regularizer rather than a major driver of accuracy.

    \subsubsection{Attention--gaze agreement}

    \begin{table*}[t]
    \centering
    \caption{Saliency evaluation on the test set. $\uparrow$ indicates higher is better, $\downarrow$ indicates lower is better.}
    \label{tab:saliency_results}
    \setlength{\tabcolsep}{4pt}
    \begin{tabular}{llccccccc}
    \toprule
    Backbone & Method & AUC$\uparrow$ & NSS$\uparrow$ & CC$\uparrow$ & EMD$\downarrow$ & SIM$\uparrow$ & KL$\downarrow$ & IG$\uparrow$ \\
    \midrule
    \multirow{5}{*}{DINOv3} 
    & Baseline & 0.826 & 1.263 & 0.427 & 26.640 & 0.400 & 1.266 & 0.898 \\
    & GII injection & 0.803 & 1.108 & 0.379 & 29.036 & 0.376 & 1.358 & 0.750 \\
    & EGViT & 0.813 & 1.199 & 0.414 & 27.433 & 0.382 & 1.321 & 0.784 \\
    & EG-PCS-Net (Raw) & \textbf{0.898} & \textbf{1.718} & \textbf{0.585} & 18.594 & \textbf{0.498} & \textbf{0.875} & \textbf{1.494} \\
    & EG-PCS-Net (Rollout) & 0.895 & 1.702 & 0.580 & \textbf{18.472} & 0.497 & 0.897 & 1.470 \\
    \midrule
    \multirow{5}{*}{DeiT III} 
    & Baseline & 0.717 & 0.007 & 0.002 & 46.692 & 0.201 & 2.151 & -0.507 \\
    & GII injection & 0.722 & 0.026 & 0.009 & 48.195 & 0.210 & 2.154 & -0.489 \\
    & EGViT & 0.724 & 0.036 & 0.013 & 49.942 & 0.210 & 2.144 & -0.503 \\
    & EG-PCS-Net (Raw) & \textbf{0.884} & \textbf{1.514} & \textbf{0.563} & 15.119 & 0.480 & \textbf{0.945} & \textbf{1.253} \\
    & EG-PCS-Net (Rollout) & 0.880 & 1.501 & 0.558 & \textbf{14.806} & \textbf{0.483} & 0.953 & 1.230 \\
    \midrule
    \multirow{5}{*}{CLIP} 
    & Baseline & 0.700 & 0.192 & 0.071 & 33.670 & 0.242 & 1.953 & -0.343 \\
    & GII injection & 0.707 & 0.199 & 0.079 & 33.018 & 0.267 & 1.888 & -0.315 \\
    & EGViT & 0.736 & 0.334 & 0.128 & 32.642 & 0.284 & 1.843 & -0.182 \\
    & EG-PCS-Net (Raw) & \textbf{0.874} & \textbf{1.423} & \textbf{0.551} & \textbf{15.283} & \textbf{0.497} & \textbf{0.929} & \textbf{1.159} \\
    & EG-PCS-Net (Rollout) & 0.872 & 1.415 & 0.541 & 15.666 & 0.487 & 0.990 & 1.119 \\
    \bottomrule
    \end{tabular}
    \end{table*}
    
    The attention--gaze agreement results in Table~\ref{tab:saliency_results} are computed on the gaze-compatible portion of the test split. For each backbone and training method, we select the best checkpoint (based on validation accuracy across 10 runs) and extract model attention maps using the \emph{Raw} and \emph{Rollout} methods. These attention maps are directly compared to the corresponding gaze saliency maps for every image in the test split using standard saliency metrics (AUC, NSS, CC, EMD, SIM, KL, IG). Each metric is computed for every test image and then averaged across the full test set. For each metric, we report the stronger averaged result obtained between the Raw and Rollout extraction methods.
        
    Under this evaluation protocol, EG-PCS-Net consistently outperforms the alternative gaze-integration strategies across backbones. It achieves higher AUC, NSS, CC, SIM, and IG while simultaneously reducing EMD and KL divergence, indicating that its attention maps both better predict fixation locations and more closely match the spatial distribution of human gaze. The improvements are particularly pronounced for DeiT III and CLIP, where the baseline models show near-zero (and sometimes negative) correlation values, but EG-PCS-Net shifts these metrics into strong positive agreement. For DINOv3, which already shows comparatively stronger agreement without gaze supervision, gains are smaller but remain consistent across metrics.
    
    Taken together, these results show that training with gaze makes the model look at images more like humans do. While predictive accuracy remains similar, the attention maps produced by EG-PCS-Net are more consistent with where people actually fixate when judging cycling safety. In practical terms, the model not only predicts which scene looks safer, but also focuses on visual regions that humans consider relevant for that judgment.
        
    \subsection{Qualitative Analysis}

    \begin{figure*}[t]
        \centering
        \includegraphics[width=0.88\linewidth]{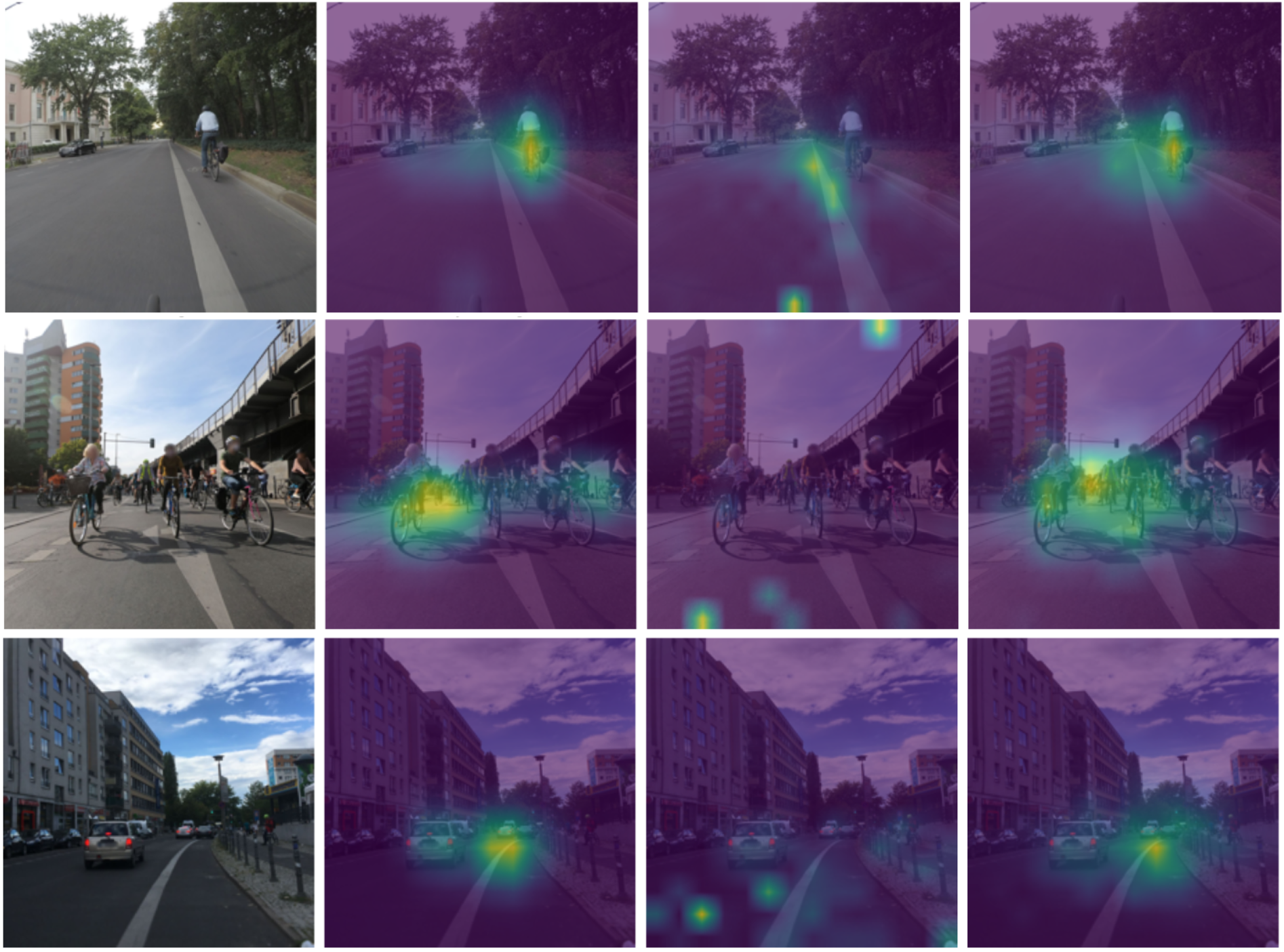}
        \caption{Qualitative comparison examples. From top to bottom: DINOv3, DeiT III, and CLIP. In each row (left-to-right): input image, eye-tracking gaze saliency, baseline attention for the corresponding backbone, and EG-PCS-Net attention (ours). For DINOv3, the baseline attention is concentrated along the lane boundary separating the cycling lane, whereas EG-PCS-Net shifts the focus toward the cyclist, aligning more closely with the human gaze pattern. For DeiT III, the baseline attention is diffuse and weakly structured, while EG-PCS-Net shifts attention toward salient traffic participants. For CLIP, baseline attention is scattered across less relevant regions, whereas EG-PCS-Net concentrates attention on human-relevant areas such as the car and the cycling lane, yielding visibly stronger agreement with gaze.}
        \label{fig:qual_all}
    \end{figure*}

    Figure ~\ref{fig:qual_all} illustrate three representative scenarios comparing (i) the input image, (ii) the eye-tracker gaze saliency, (iii) ViT attention from a model trained without gaze (Baseline), and (iv) ViT attention from EG-PCS-Net. In each example, EG-PCS-Net concentrates attention on regions that better match human fixations (e.g., nearby vehicles, cyclists, and the relevant roadway corridor), while the baseline model attends to less informative or exhibits more diffuse focus.

    Consistent with the quantitative results, the improvement is most visually apparent for DeiT III and CLIP, where many of the most gaze-aligned examples under EG-PCS-Net were not well-aligned under the baseline model. For DINOv3, qualitative differences are typically smaller: many examples already show reasonable alignment without gaze-guided training, and EG-PCS-Net mainly refines attention concentration rather than shifting it to entirely new regions. This matches the interpretation that strong self-supervised pretraining in DINOv3 produces attention patterns that are already partially consistent with the task-relevant human viewing behavior.
    
    \section{Conclusions and Limitations}
    
    This work explores incorporating human gaze into pairwise cycling-safety modeling to improve both predictive performance and attention interpretability. Across three pretrained transformer backbones, gaze integration yielded modest but consistent predictive benefits, demonstrating that human visual signals can act as a useful supervisory cue. More importantly, our proposed attention-alignment strategy substantially improved agreement between model attention and human fixation patterns, clearly outperforming alternative gaze-integration state of the art methods . While competing strategies introduced gaze during training, they did not consistently translate into improved attention–gaze alignment. In contrast, EG-PCS-Net achieved large gains in alignment metrics while maintaining competitive classification and ranking accuracy, showing that explicitly supervising attention is an effective way to bridge model reasoning and human visual behavior.
    
    Some limitations should be acknowledged. First, the gaze-annotated subset used in this study is considerably smaller than those employed in prior gaze-guided transformer works \cite{Ma2023EGViT,Chen2026GIIViT}, where gaze annotations were available for the full training dataset. Second, our implementation differs from those studies in both backbone selection and attention mechanism: we rely on different pretrained transformers and use their intrinsic self-attention for alignment, whereas prior works adopted different architectures and Grad-CAM–based attention. These differences limit strict comparability between alignment results across studies.

    \section*{Acknowledgements}
    This work is funded by LARSyS funding (DOI: \nolinkurl{10.54499/LA/P/0083/2020}, \nolinkurl{10.54499/UIDP/50009/2020}, and \nolinkurl{10.54499/UIDB/50009/2020}), through Fundação para a Ciência e a Tecnologia, and the Department of Technology, Management, and Economics at the Technical University of Denmark (DTU). C. Santiago and M. Marques are also supported by the PT Smart Retail project (PRR - \nolinkurl{02/C05-i11/2024.C645440011-00000062}), through IAPMEI - Agência para a Competitividade e Inovação.

	\bibliographystyle{IEEEtran}
	\bibliography{root} 
	
\end{document}